\documentclass[letterpaper, 10 pt, conference]{ieeeconf}  % Comment this line out if you need a4paper

\IEEEoverridecommandlockouts                              % This command is only needed if 
\usepackage{subcaption}
\usepackage[table]{xcolor}
\usepackage{multirow}
\usepackage{graphicx}
\usepackage{amsmath}

\usepackage{hyperref}
\title{\LARGE \bf
Bio-Inspired Plastic Neural Networks for Zero-Shot Out-of-Distribution Generalization in Complex Animal-Inspired Robots
}

\author{Binggwong Leung$^{1, \dagger}$, Worasuchad Haomachai$^{1, \dagger}$, Joachim Winther Pedersen$^{2}$, \\Sebastian Risi$^{2, *}$, and Poramate Manoonpong$^{1, 3, *}$% <-this % stops a space
\thanks{*Corresponding authors}% <-this % stops a space
\thanks{$^\dagger$These authors contributed equally to this work}
\thanks{This work is supported by the Startup Grant on Bio-inspired Robotics and a Student Research Grant from the Vidyasirimedhi Institute of Science and Technology (VISTEC), the European Union (ERC, GROW-AI, 101045094), and a Sapere Aude: DFF Starting Grant (9063-00046B).}% <-this % stops a space
\thanks{$^{1}$Binggwong Leung, Worasuchad Haomachai and Poramate Manoonpong are with Bio-inspired Robotics and Neural Engineering Lab,
School of Information Science and Technology (IST), Vidyasirimedhi Institute of Science and Technology, 21210 Rayong,  Thailand,
        {\tt\small binggwong.l\_s17@vistec.ac.th, haomachai@gmail.com, poramate.m@vistec.ac.th}}%
\thanks{$^{2}$Joachim Winther Pedersen and Sebastian Risi are with the IT University of Copenhagen,
        2300 Copenhagen, Denmark,
        {\tt\small jwin@itu.dk, sebr@itu.dk}}%
\thanks{$^{3}$Poramate Manoonpong is also with the Embodied Artificial Intelligence and Neurorobotics Laboratory, SDU Biorobotics, The Marsk Mc-Kinney Moller Institute, University of Southern Denmark, Odense, Denmark.
}
}

\begin{document}

\maketitle
\thispagestyle{plain}
\pagestyle{plain}

%%%%%%%%%%%%%%%%%%%%%%%%%%%%%%%%%%%%%%%%%%%%%%%%%%%%%%%%%%%%%%%%%%%%%%%%%%%%%%%%
\begin{abstract}
Artificial neural networks can be used to solve a variety of robotic tasks. However, they risk failing catastrophically when faced with out-of-distribution (OOD) situations. Several approaches have employed a type of synaptic plasticity known as Hebbian learning that can dynamically adjust weights based on local neural activities. Research has shown that synaptic plasticity can make policies more robust and help them adapt to unforeseen changes in the environment. However, networks augmented with Hebbian learning can lead to weight divergence, resulting in network instability. Furthermore, such Hebbian networks have not yet been applied to solve legged locomotion in complex real robots with many degrees of freedom. In this work, we improve the Hebbian network with a weight normalization mechanism for preventing weight divergence, analyze the principal components of the Hebbian's weights, and perform a thorough evaluation of network performance in locomotion control for real 18-DOF dung beetle-like and 16-DOF gecko-like robots. We find that the Hebbian-based plastic network can execute zero-shot sim-to-real adaptation locomotion and generalize to unseen conditions, such as uneven terrain and morphological damage. %Compared to feedforward networks without plasticity and LSTM networks, the Hebbian network exhibits generalization capabilities in environments not seen during training.  

\end{abstract}

%%%%%%%%%%%%%%%%%%%%%%%%%%%%%%%%%%%%%%%%%%%%%%%%%%%%%%%%%%%%%%%%%%%%%%%%%%%%%%%%
\section{Introduction}

In the field of machine learning research, deep neural networks (DNNs) have been shown to be useful across a wide range of tasks \cite{goodfellow2016deep,vinyals2019grandmaster}, including robotics \cite{karoly2020deep, mouha2021deep, soori2023artificial}. However, policies for agent control based on deep neural networks tend to be brittle \cite{heaven2019deep}, meaning that they are at risk of catastrophic failure when faced with out-of-distribution (OOD) situations \cite{zhang2018study, zhao2019investigating}. A classic OOD challenge arises when the DNN controlling the robot is optimized to control a simulated robot and afterward needs to control a real-world counterpart. This problem is known as the sim-to-real gap \cite{zhao2020sim}. OOD challenges can also present themselves in the form of varying (unseen) terrains, e.g., going from a (seen) flat floor to an (unseen) uneven surface. An inability to adapt to unseen situations greatly limits the applications for robots required to navigate in the real world with all of its unforeseeable complexities. 

Approaches to mitigating these consist of extending the training set to include an abundance of random slightly different scenarios (domain randomization) \cite{Domain_randomization}, and/or developing architectures designed by hand specifically to overcome novel situations \cite{Self-Organized_Stick_Insect-Like_Locomotion, Leung_Integrated_Modular}. The disadvantages of these types of solutions are that they extend the necessary training time or risk, resulting in an architecture that is overly specific to the task for which it was designed \cite{Self-Organized_Stick_Insect-Like_Locomotion, Leung_Integrated_Modular}. 

Animals, on the other hand, demonstrate remarkable adaptability in adjusting their motor patterns to accomplish various tasks. Synaptic plasticity is thought to play an important role in supporting the adaptive capabilities of animals \cite{Citri2008}. A prominent form of synaptic plasticity, Hebbian learning, often summarized as "neurons that fire together, wire together", has been documented as a driver of synaptic change in both mammals \cite{pike1999postsynaptic, meliza2006receptive} and insects \cite{cassenaer2007hebbian}.
Recently, different formulations of Hebbian learning have been implemented for the control of artificial agents \cite{najarro2020, pedersen2021evolving, ferigo2021evolving, chalvidal2022meta, palm2021testing}. While several of these studies include experiments involving the locomotion of simulated robots, there are to the best of our knowledge not yet examples that involve complex real-world robots. The few works that demonstrated real-world transfer evolved synaptic plasticity for much simpler wheeled robots with a few degrees of freedom (DOFs) \cite{Floreano2001}.

\begin{figure}[t]
\centering
\includegraphics[width=0.49\textwidth]{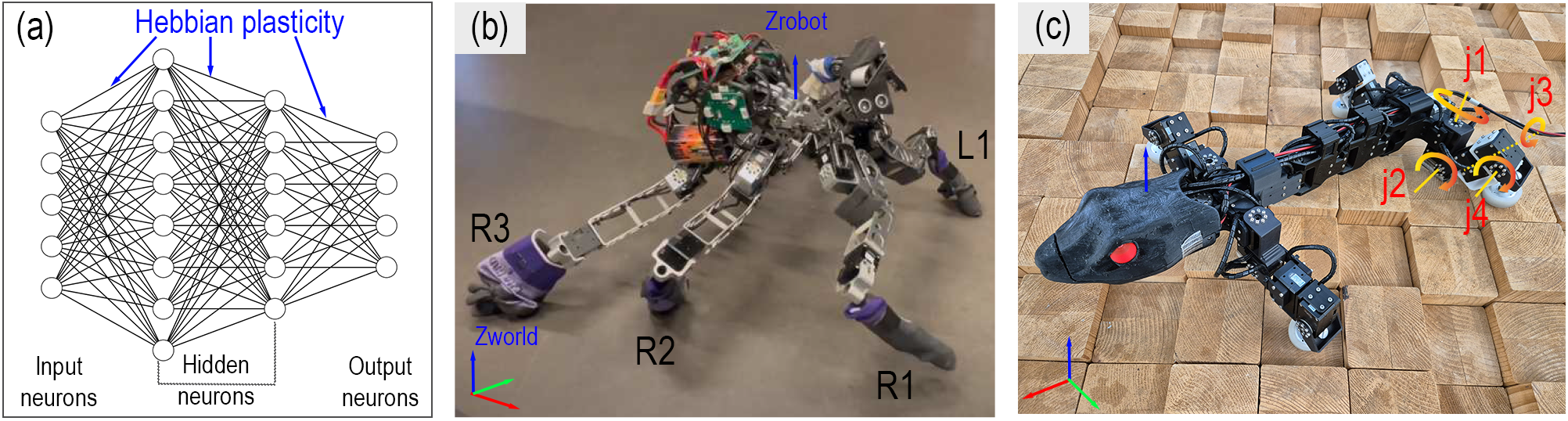}
\caption{A neural network with Hebbian plasticity (a) is trained to control a robot in simulation and then transferred to a physical robot. A dung beetle (b) and a gecko-like robot (c) were used as experimental complex platforms in this study.}
\label{fig:concept}
\end{figure}

Therefore, in this work, we demonstrate the generalization capability of a bio-inspired plastic neural network with Hebbian learning (called Hebbian network, Fig.~\ref{fig:concept}a) to be used on various real-legged robot platforms, i.e., a complex dung beetle-like robot with 18 DOFs \cite{Billeschou2020_ALPHARobot} and a gecko-like robot with 16 DOFs \cite{haomachai2024transition} (Figs.~\ref{fig:concept}b and~\ref{fig:concept}c, respectively). We first present a comparison between the Hebbian network, a simple (non-plastic) feedforward network (FF network), and a long short-term memory network (LSTM network \cite{hochreiter1997long}).
The LSTM is chosen for comparison as it is also capable of allowing the agent/robot to adapt to different environments \cite{stanley2003evolving,lstmsnake_Ouyang_2020}. We employ evolutionary strategies (ES) to optimize the weights of the FF and LSTM networks as well as Hebbian learning rule parameters. 
The first investigation and comparison utilize a dung beetle-like robot for training and evaluating of the network model.
We further examine the generalization of the Hebbian network by training a gecko-like robot to walk on flat terrain and evaluating its performance on uneven terrain and under morphological damage, i.e., one to two damaged legs.
% The results show that the robot can adapt to walk on uneven terrain while also can walk even if it encounters morphological damage, i.e., one to two of its legs are broken. 

The contribution of this study are summarized as follow: i) we present a Hebbian-based plastic network capable of zero-shot sim-to-real adaptation in locomotion and generalization to out-of-distribution scenarios for complex legged robots, without employing randomization of terrain, mass distribution, joint properties, contact friction, or morphological damage, ii) we highlight the importance of normalizing weights updated by Hebbian rules and compared standardization with max normalization to identify the most effective approach for controlling complex legged robots, iii) to understand how Hebbian plasticity enables adaptability, we analyze the weight dynamics of the robots walking under different scenarios using principal component analysis (PCA).

% Through local learning rules alone, the neural network can enable the dung beetle-like robot to perform zero-shot sim-to-real adaptation for locomotion tasks without the use of domain randomization \textcolor{red}{(i.e., terrain, mass distribution, joint properties, and contact friction randomization)} during simulation. In contrast, the LSTM performs well in the training tasks but is not as capable of generalizing from the simulation to the real world. \textcolor{blue}{As will be discussed in more detail, this points toward a reliance on domain randomization for the LSTM networks for good generalization, which is not as pronounced for the Hebbian networks.}

\section{Related Work}
\label{Related Work}

%\subsection{Legged Loco-manipulation of Animals and Robots}

%Loco-manipulation requires a combination of locomotion and object manipulation abilities to solve. A dung beetle is an example of an insect species that performs this task on a daily basis, which is why it was chosen as the model organism in this paper. Typically, a dung beetle can roll a large ball and transport a small dung pallet across terrains \cite{Leung_Integrated_Modular,Leung2020_ballrollingrule}. Varieties in the shape of dung pallets require the dung beetle to alter its movement for effective transportation. Several approaches have been inspired by the dung beetle to address various loco-manipulation tasks \cite{gong2023legged, Thor2018, Modular_Neural_Control_Chris,leung2018_modulaarNeuralControl}. These methods generally employ feedforward networks without synaptic plasticity to solve specific tasks.

\subsection{Hebbian Learning}
Synaptic plasticity is nearly ubiquitous in biological brains \cite{Citri2008}. Since the ability of biological synapses to change over time is thought to underlie much of the great behavioral adaptiveness found in animals, there has long been an interest in implementing similar plasticity into artificial neural networks (ANNs) \cite{soltoggio2018born,risi:book25}. 
Plastic neural networks differ from most standard ANNs in that the main focus of optimization during the training phase is not on the synaptic weight values, but on the parameters that will change these weight values over time in response to inputs to the network. 

Several different ways of implementing synaptic plasticity in ANNs have been proposed. Soltoggio et al. \cite{soltoggio2008evolutionary} evolved a neuromodulated plastic network to solve the T-maze task with non-stationary rewards. This approach was later used in combination with a deep autoencoder for feature extraction to solve more elaborate tasks, such as Configurable Tree Graphs as well as a double T-maze constructed within the Malmo Minecraft Environment \cite{ben2020evolving}.  Adaptive HyperNEAT by Risi and Stanley \cite{risi2010indirectly} indirectly encoded the parameters of local learning rules and used them to update the weights of a neural network in a regular, patterned manner to solve a T-maze task.
More recently, Najarro and Risi \cite{najarro2020} evolved a simple synaptic plasticity rule for each synapse in an ANN to control the locomotion of a simulated four-legged robot. This approach demonstrated the possibility of rapidly transforming a randomly initialized network into a functional one and adapting to leg damages not seen during the evolution of the plasticity rules. Since this approach has been shown to work for locomotion in simulation \cite{najarro2020, pedersen2021evolving}, the approach presented here is building on it for the sim-to-real locomotion adaption of complex real-world robots.

\subsection{Robots with Adaptive Locomotion}
Several efforts have been made to make the locomotion of robots more adaptive. Cully et al. \cite{cully2015robots} used quality diversity algorithms to learn a repertoire of different gaits. In novel circumstances, Bayesian Optimization can be used to select the most optimal gait in its repertoire.
Yang et al. \cite{MELA_quadruped} optimized a policy that can dynamically combine the outputs of several neural networks, each pre-trained to produce a specialized skill. Larsen et al. \cite{Self-Organized_Stick_Insect-Like_Locomotion} used decentralized neural pattern generation in a nature-inspired approach based on stick insect locomotion, enabling self-organized locomotion for a stick insect-like robot. 

Kumar et al.~\cite{kumar2021rma} took a different approach to handling OOD scenarios. Their technique, Rapid Motor Adaptation, utilized two different training phases in simulation. In the first phase, a policy network was optimized with observation inputs and information about the current configuration of the environmental setup, such as the level of friction, payload, and motor strengths. In the second phase, a different network was optimized in a supervised manner to predict the environmental information and pass it to the policy. The optimization was performed across several environmental configurations, including different levels of rugged terrain. Wang et al. \cite{wang2023learning} extended this approach by adding a term to the reward function that encourages the learning of natural gaits using motion capture data from a real-world dog \cite{zhang2018mode}.
Extending the training data by randomizing settings has been commonly used to make policies more robust to unseen OOD situations \cite{tobin2017domain, wang2022generalizing, muratore2022robot}. Peng et al. \cite{peng2018sim} demonstrated this in experiments involving a robotic arm. Randomization of environmental configurations also played an important role in the more intricate robot manipulation task of solving a Rubik's Cube with a humanoid hand, as accomplished by Akkaya et al. \cite{akkaya2019solving}. 

These approaches all involve an elaborate architecture, training scheme, and/or extension of the training data with randomized configurations. In contrast, the approach of evolving synaptic plasticity for sim-to-real adaptation and robust policies proposed here is conceptually simple and does not require a randomization of terrain, mass distribution, joint properties, contact friction, or morphological damage to extend the training data for sim-to-real locomotion adaptation.

\section{Materials and Methods}

\subsection{Bio-inspired Plastic Neural Networks}

For the bio-inspired network, we employ a feedforward network incorporating a synaptic plasticity mechanism inspired by Hebbian plasticity \cite{najarro2020,soltoggio2018born,soltoggio2008evolutionary}. The ABCD Hebbian learning rule is utilized to update the synaptic weights between neurons in the feedforward network. The following equation describes the weight update process based on the pre-synaptic and post-synaptic activities of neurons.
\begin{equation}
    \Delta w_{ij}^k = \eta\cdot({A o_i o_j + Bo_i + Co_j + D}),
\end{equation}
where $w_{ij}^k$ represents the weight between neurons \emph{i} and \emph{j}. $k$ is a number representing each layer in the neural network. $\eta$ is the evolved learning rate. $A$ is an evolved correlation term. $B$ is an evolved pre-synaptic term. $C$ is an evolved post-synaptic term, with $o_i$ and $o_j$ being the pre-synaptic and post-synaptic activations, respectively. The coefficients $A$, $B$, and $C$ explicitly determine the local dynamics of the network weights based on neural activations and the evolved coefficient $D$ is a bias term. Following Najarro and Risi~\cite{najarro2020}, a separate ABCD-rule is optimized for each weight in the network, and the weight values are updated each time the neurons are activated.

\subsection{Weight Normalization}

% After updating, the weights in each layer are normalized.
Weight normalization is crucial to improve the network stability by preventing the weights from becoming excessively large (i.e., preventing divergence) as the Hebbian updates are continually added to them. In this work, we compare two simple but effective normalization methods to identify the optimal approach for controlling complex legged robots. First, the weights are normalized layer-wise by dividing them by the maximum weight of that layer (max normalization), according to the following equation:
\begin{equation}
\label{eq:maxnorm}
    w_{ij}^k = \frac{w_{ij}^k}{w_{max}^k} = \frac{w_{ij}^k}{max(abs(W_k))},
\end{equation}
where $k$ is a number representing each layer in the neural network. The matrix $W_k$ contains the weights of layer $k$.  
Second, a standard deviation normalization method is applied \cite{najarro2020, pedersen2021evolving}, as shown in the following equation:
\begin{equation}
\label{eq:varnorm}
    w_{ij}^k = \frac{w_{ij}^k-\overline W_k}{\sigma_{W_k}},
\end{equation}
where $\overline W_k$ is the mean of the weights of layer $k$. $\sigma_{W_k}$ is the standard deviation of the weights of layer $k$. The comparison results are presented in the Results section.

\section{Experiments}
\label{experiments}
% describe 

\subsection{Robot Locomotion Training}

As our first investigation, we trained a six-legged dung beetle-like robot in a simulated environment using the Omniverse Isaac Gym Reinforcement Learning Environments \cite{DBLP:conf/corl/RudinHR021}.
Each leg of the robot has three actuated joints (Fig.~\ref{fig:concept}b). The ANN receives sensory inputs, including 18 joint angles, six foot contacts, and three body orientation values (roll, pitch, yaw). The ANN generates joint position commands for the robot’s motors. Detailed information on the robot’s motor output and sensory feedback is provided in Table \ref{tab:my-table1} (see \cite{Billeschou2020_ALPHARobot} for further details of the robot).
After training in simulation, the learned policies were evaluated on flat terrain and then tested on the real-world robot to assess their generalization across the sim-to-real gap.

% We simulated multiple robots in parallel using the Omniverse Isaac Gym Reinforcement Learning Environments \cite{DBLP:conf/corl/RudinHR021}. 
% As our first investigation, the six-legged dung beetle-like robot was used in the simulation. Each leg of the robot has three actuated joints (Fig.~\ref{fig:concept}b). The ANN receives inputs from the robot in the form of joint angles, body orientation (roll, pitch, yaw), and foot contact feedback. The ANN's outputs represent the joint position commands for the robots' motors. Information on the robot's motor output and sensory feedback is displayed in Table \ref{tab:my-table1}; See \cite{Billeschou2020_ALPHARobot} for further details of the robot.

% Please add the following required packages to your document preamble:
% \usepackage{multirow}
\begin{table}[t]
\centering
\caption{Robot motor outputs and sensory feedback.}
\label{tab:my-table1}
\begin{tabular}{|c|l|l|}
\hline
\rowcolor{gray!30}
                       & \textbf{Type}             & \textbf{Description}                                                                                  \\ \hline
\multirow{3}{*}{\shortstack[l]{Sensory\\ feedback}} & Joint angle      & range: [-1, 1] radian.                                                                      \\ \cline{2-3} 
                       & Foot contact     & \begin{tabular}[c]{@{}l@{}} 0 : No contact, \\ 1 : Leg is in contact.\end{tabular} \\ \cline{2-3} 
                       & Body orientation & Roll, pitch, yaw angles; \\ & & range: [-3.14, 3.14] radian.                                                           \\ \hline
Motor output & Joint angle & range: [-1, 1] radian.                                                                     \\ \hline
\end{tabular}
\end{table}

% \subsection{Robot Locomotion Control}
% We evaluated the trained policies on their ability to let the dung beetle-like robot move forward on flat terrain. The ANN receives all sensory inputs from the robot and generates outputs that control all legs. The ANN receives 18 joint angles, six foot contacts, and three body orientation values as inputs. The optimized policies are then evaluated on the real-world robot to test whether the policy can generalize across the sim-to-real gap.

\subsection{Optimization Details}
\label{optimization algorithm}

This study uses evolution strategies (ES) to optimize neural network weights.
% ES has the advantage of not requiring the gradient for backpropagation to be calculated and can handle both dense and sparse rewards \cite{salimans2017evolution}. 
ES has the advantage of being gradient-free, eliminating the need for backpropagation while effectively handling both dense and sparse rewards  \cite{salimans2017evolution}.
ES has achieved comparable performance to the proximal policy optimization (PPO) approach in robotics \cite{Efficacy_of_Modern_Neuro-Evolutionary}.
Pagliuca et al. \cite{Efficacy_of_Modern_Neuro-Evolutionary} have shown that reward functions optimized for ES do not perform effectively with PPO, and vice versa. This suggests that a fair comparison between algorithms from different classes should involve reward functions specifically optimized for each algorithm. Otherwise, comparisons based on reward functions tailored to one method may introduce bias.
In this study, we design the reward function (described  next) specifically for ES. Therefore, a direct comparison with PPO would require optimizing a separate reward function for PPO, which is beyond the scope of this work.

At the beginning of the optimization process, the parameters of the neural network are initialized. At each generation or evolutionary step $t$, the parameters are tested on an agent, and the average fitness $F(h_t)$ of $n$ agents is calculated using the reward function. A new set of parameters $h_{t+1}$ is then created by adding a normal noise ($\epsilon$) to the parameters in the previous step $h_{t}$, using the following equation:
\begin{equation}
    \boldsymbol{h_{t+1}}=\boldsymbol{h_t}+\frac{\alpha}{n\sigma} \sum_{i=1}^n F_i \cdot (\boldsymbol{h_t}+\sigma\epsilon_i),
\end{equation}
where $\alpha$ determines the step size for updating the parameters at each generation, and $\sigma$ defines the amount of noise introduced to the parameters at each generation. $\alpha$ and $\sigma$ are initialized as 0.1, and the decay rate per generation of both values is 0.999. We used a population size of 1,024 individuals ($n$) over 500 generations.

The ES is used to optimize the trainable parameters for all models. The weights for the FF network are initialized uniformly with $w\in(-0.1, 0.1)$. The weights of the Hebbian network are initialized uniformly with $w\in(-0.01, 0.01)$, whereas the Hebbian coefficients are initialized with a normal distribution with a mean of zero and a standard deviation of 0.01. The weights of the LSTM network are initialized with $w\in(-0.1, 0.1)$, whereas the hidden and cell states are initialized with $w\in(-0.01, 0.01)$. These initializations were found through hand-tuning in preliminary experiments.
% $\eta_i=N(0,0.01)$ to the hebbian coefficients
\subsection{Reward Function}
\label{Rewards function}

The robot's reward is taking into account the body velocities and body orientation based on the following formula:

\begin{equation}
\label{eq:reward_fucntion}
    R = \sum_{t=0}^n (k_v V_x(t) + k_u U(t) + k_{yaw} Yaw(t)),
\end{equation}
\begin{equation}
  U(t) =
    \begin{cases}
      0 & \text{$proj_{\hat{z}_{world}} ~z_{robot}$ $>$ 0.93},\\
      -0.5 & \text{otherwise},\\
    \end{cases}
\end{equation}
\begin{equation}
  Yaw(t) =
    \begin{cases}
      0 & \text{if $abs(yaw)$ $<$ 0.45 radian},\\
      -0.5 & \text{otherwise},\\
    \end{cases}
\end{equation}
where $R$ is the accumulative reward over simulation step $t$. The robot performs $n$ ($n$=500) simulation steps. $k_v, k_u, k_y$ represent the coefficients for each reward term, equating to 2.0, 0.5, and 0.5, respectively. $V_x$ is the robot's velocity in the x-axis of the world frame. $U(t)$ is an upright posture reward calculated from the projection of the robot's z-axis vector ($\hat{z}_{robot}$) to the world's z-axis ($\hat{z}_{world}$) (Fig.~\ref{fig:concept}b). $Yaw(t)$ is the heading direction reward calculated from the yaw angle of the robot. 
% The rewards accumulate over $n$ simulation steps for each episode. 
% For the locomotion task, the reward is calculated as Eq.~\ref{eq:reward_fucntion}.

\subsection{Model Details}
\label{Model details}
% LSTM
% Architectures
We use (1) a feedforward network (FF network), (2) a feedforward network with Hebbian plasticity (Hebbian network), and (3) a long short-term memory network (LSTM network) for training locomotion. These neural networks has 27 inputs and 18 outputs.
% The inputs of all models vary depending on the task. For this locomotion task, the neural network receives 27 inputs and generates 18 outputs.

FF and Hebbian networks have the same feedforward architecture with 64 and 32 neurons in the hidden layers. However, the Hebbian networks update the weights of the model using the optimised Hebbian learning rule. The total number of weights of the FF network is 4,352  (layer sizes: 27, 64, 32, 18). The Hebbian networks contain five times more parameters than the FF network to optimize due to the Hebbian coefficients ($A, B, C, D$) and learning rates ($\eta$) to update each weight individually. Thus, the Hebbian networks have 21,760 parameters.

The LSTM network has a hidden state size of 60. The total number of the weights of the LSTM network is  22,686 (layer sizes: 27, 60, 18). The number of elements in the hidden states and cell states of the LSTM network is 60 (Table~\ref{tab:my-table_Number_parameters}).
This network size is chosen to match the number of trainable parameters in the Hebbian network for a fair comparison. 
% We also add an LSTM network with 100 elements in its hidden state (LSTM\_big) as an additional comparison.  
%------------------------------------------

\begin{table}[h]
\centering
\caption{Number of evolved parameters (fixed after training) and plastic parameters of the models.}
\label{tab:my-table_Number_parameters}
\begin{tabular}{|l|l|l|l|}
\hline
\rowcolor{gray!30}
\textbf{ } & \multicolumn{1}{l|}{\textbf{FF}} & \textbf{Hebbian} & \textbf{LSTM}       \\ \hline
\textbf{Evolved} & \multicolumn{1}{l|}{4,352} & 21,760 & 22,686                     \\ \hline
\textbf{Plastic} & \multicolumn{1}{l|}{0} & 4,352 & 120                  \\ \hline
\end{tabular}
\end{table}

%------------------------------------------

% \begin{table}[h]
% \centering
% \caption{Number of evolved parameters (fixed after training) and plastic parameters of the models.}
% \label{tab:my-table_Number_parameters}
% \begin{tabular}{|l|ll|}
% \hline
%               % & \multicolumn{2}{l|}{\textbf{Locomotion task}}   \\ \hline
% \textbf{Name} & \multicolumn{1}{l|}{Evolved} & Plastic       \\ \hline
% \textbf{FF} & \multicolumn{1}{l|}{4352}        & 0                        \\ \hline
% \textbf{Hebbian} & \multicolumn{1}{l|}{21,760}      & 4352                   \\ \hline
% \textbf{LSTM} & \multicolumn{1}{l|}{22,686}      & 120                    \\ \hline
% \textbf{LSTM\_big} & \multicolumn{1}{l|}{53,486}      & 200                    \\ \hline

% \end{tabular}
% \end{table}

%  single LSTM 6,912 and 3,420
% seqLSTM 21,708 params:(locomtion)
% seqLSTM 10,792 params:(object)
% 
% For the LSTM, we use a single LSTM module for our tasks. There are 6,912 and 3,420 parameters in total for locomotion and object transportation tasks, respectively.

\section{Results}
\label{Results}

% \subsection{Normalization Methods}
% From the results in Fig. 2, we found that the max-normalization method provides a better result for both locomotion and object transportation tasks. The max-normalization approach gives more variation in the rewards. It also gives a solution with higher rewards for both tasks. Therefore, we use this normalization method as a standard method for experiments. We hypothesized that the max-normalization approach might limit the amplitude of the weight so that it does not grow too large, thereby leading to a better result.

% \begin{figure}[h]
%

\subsection{Dung Beetle-like Robot Locomotion}

\begin{figure}[t]
\centering
\includegraphics[width=0.40\textwidth]{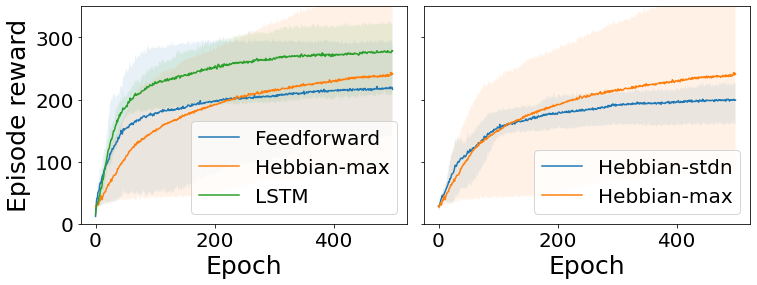}
\caption{(left) Training curves: Dung-Beetle robot locomotion.  The graph shows the average and standard deviation of the best individual's performance across five trials for each model. (right) Comparison of using standard deviation normalization (Hebbian-stdn, see Eq.~\ref{eq:varnorm}) and maximum normalization (Hebbian-max, see Eq.~\ref{eq:maxnorm}) methods to normalize the dynamical weights of the Hebbain network. Utilizing the Hebbian-max method yields a better-performing solution.}
\label{train curve locomotion}
\end{figure}

Training curves averaged over five trials look similar for the feedforward and Hebbian networks, with the LSTM-based approach having a slightly better final average performance  (Fig.~\ref{train curve locomotion}, left). 
% \textcolor{green}{TODO: why LSTM better than LSTM\_big?}

\begin{figure}[t]
\centering
\includegraphics[width=0.40\textwidth]{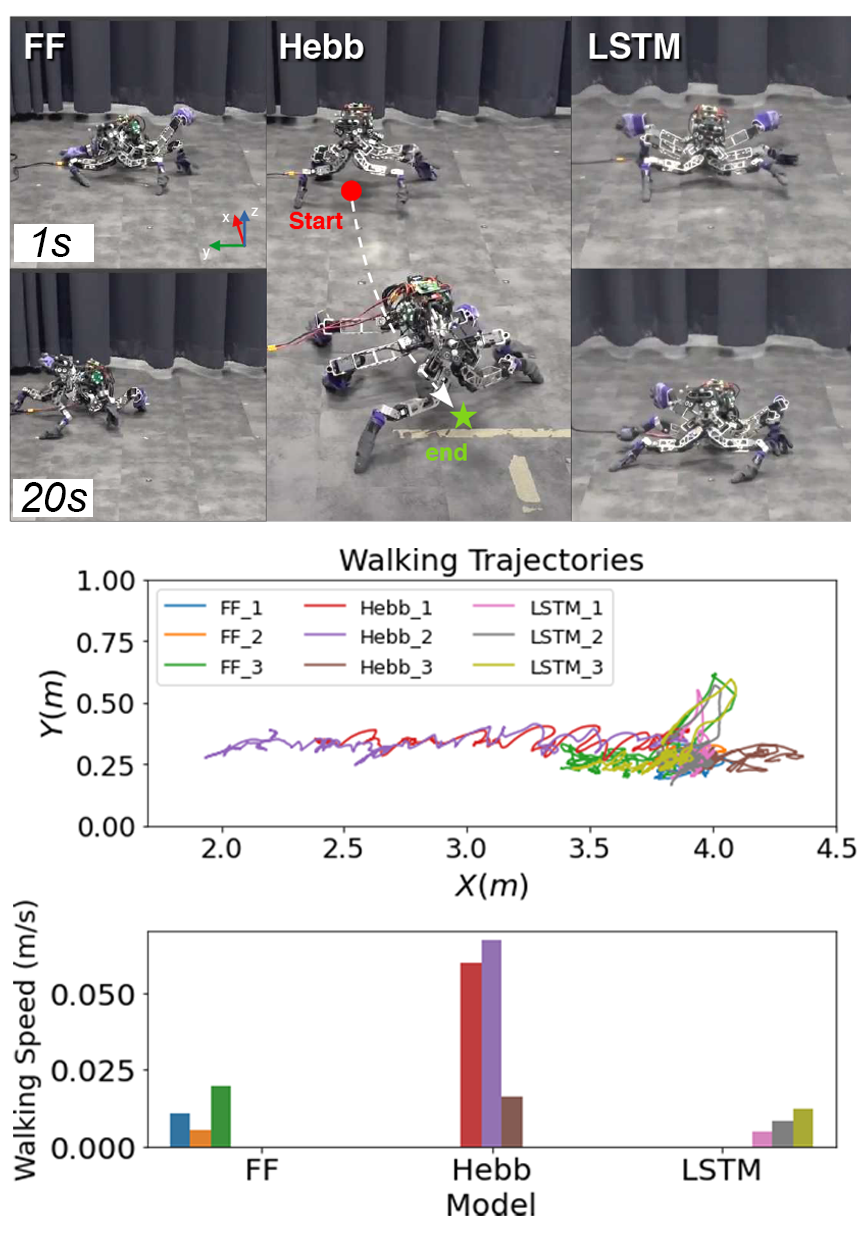}
\caption{Locomotion of the real-world dung beetle-like robot. Snapshots, walking trajectories, and walking speed (m/s) are shown. The labeled numbers (\_1, \_2, \_3) represent the ranking of the model arranged according to the training reward. Real robot locomotion can be seen at \url{https://bit.ly/3D2ZHBf}.}
\label{real robot walk}
\end{figure}

\begin{figure}[t]
\centering
\includegraphics[width=0.48\textwidth]{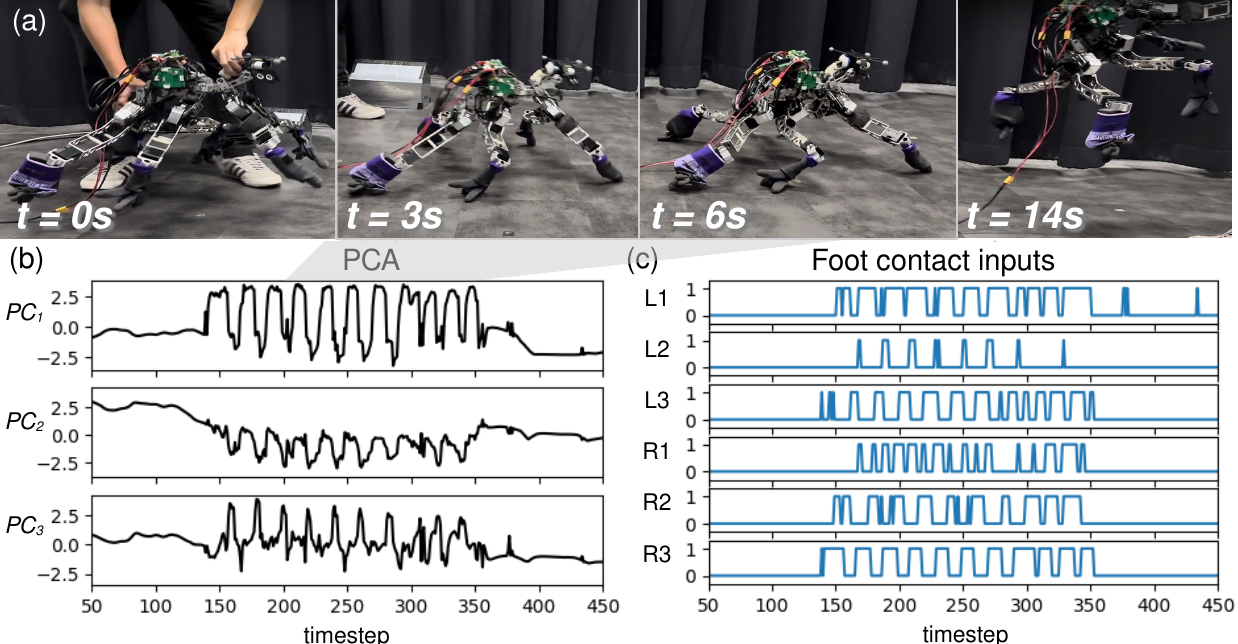}
\caption{The robot with the Hebbian network can automatically initiate walking behavior based on foot contact inputs. When placed on the ground, the robot it begins to walk. When the robot is lifted, the legs stop oscillating (see also \url{https://bit.ly/3D2ZHBf}).}
\label{fig:hebb_pick_up}
\end{figure}

We tested the top three solutions resulting from the five runs of each model on the dung beetle-like robot (Fig.~\ref{real robot walk}). In this sim-to-real adaptation test setting, the Hebbian network greatly outperformed the other models. The robot controlled by the Hebbian network (Hebb\_2) walked at the fastest speed ($\approx$7cm/s) in the experiment. The simple feedforward and LSTM policies barely moved from their starting positions in the 20 seconds during which the models were tested. The Hebbian network also revealed interesting locomotion behaviors. The robot did not move its leg until it was put on the floor. When the robot's feet made contact with the ground, the robot started walking but stopped moving again when picked up from the ground (Fig.~\ref{fig:hebb_pick_up}).

An additional experiment compared the normalization method applied to the Hebbian weight adaptation. We found that dividing by the largest absolute weight value within the network yielded the best learning performance (see Fig.~\ref{train curve locomotion}, right). Consequently, models trained using the max normalization method were selected for further investigation.

% \textcolor{red}{An additional experiment, a comparison of the normalization methods on the Hebbian weights is shown in Fig.~\ref{train curve locomotion}. As dividing with the largest absolute weight value in the network yielded the best learning performance, models trained with this method were used for the analysis of the results.}

% As an additional experiment, a comparison of the normalization methods on the Hebbian weights is shown in Fig.~\ref{compare normalization}. As dividing with the largest absolute weight value in the network yielded the best results in both tasks, models trained with this method were used for the analysis of the results.

\subsection{Hebbian Network Analysis}

\begin{figure}[t]
\centering
\includegraphics[width=0.49\textwidth]{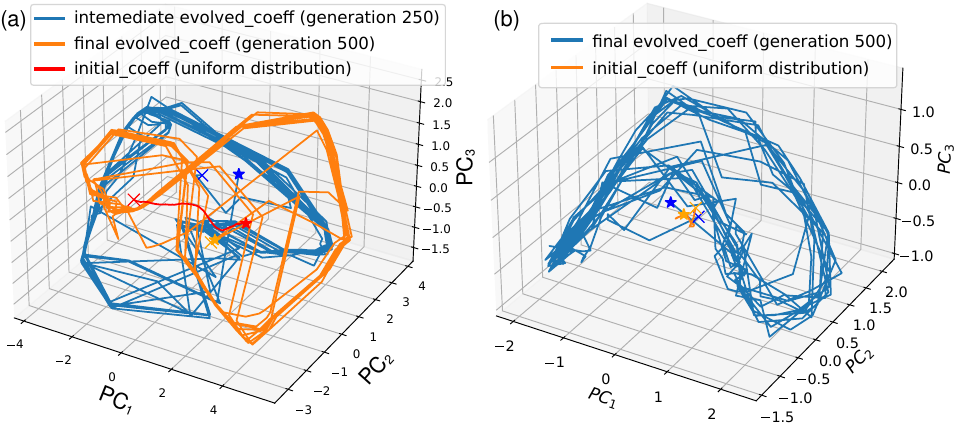}
\caption{Trajectory of the plastic weights. (a) Principal component analysis (PCA) on the weight space of the Hebbian network optimized on the locomotion task in the robot simulation. (b) PCA on the hidden states of the LSTM network on the locomotion task in robot simulation. Only the LSTM network with trained parameters exhibits a limit cycle in the PCs of the hidden states. The 'x' and star symbols indicate the value at the beginning and end of the episode, respectively.}
\label{PCA analysis Hebb}
\end{figure}

The locomotion experiment revealed that the Hebbian network can generate rhythmic motor outputs to allow the legged robot to walk. To investigate whether the rhythmic outputs of the model may be caused by the oscillation of synaptic weights, we used principal component analysis (PCA) on the weights of the neural network to visualize potential oscillation patterns. In Fig. \ref{PCA analysis Hebb}, a limit cycle attractor can be observed in the weights of the ANN with Hebbian updates. 
% \textcolor{blue}{For non-optimized Hebbian rules, no limit cycle was present, and the weights were continually updated toward a single direction in the weight space.}
In contrast, non-optimized Hebbian rules cannot achieve a limit cycle attractor and the weights were continually updated toward a single direction in the weight space (fixed point attractor).

\subsection{Gecko-like Robot Locomotion}

We further investigated the use of a Hebbian network \footnote{Note that the network architecture for the gecko-like robot is identical to that used for the dung beetle-like robot. Both networks share the same number of layers and reward function. However, the input joint angles and output motor neurons are adjusted to match the specific number of joints in each robot.} for locomotion control in another high-degree-of-freedom robot. In this case, a gecko-like robot was used (Fig.~\ref{fig:concept}c) \cite{haomachai2024transition}.
% gecko robot configuration
Its structure comprises four identical legs, each consisting of four joints for leg forward/backward motion (j1), leg elevation/depression (j2), foot attachment/detachment (j3), and leg flexion/extension (j4). In total, the gecko-like robot has 16 degrees of freedom (DOFs).
Therefore, the Hebbian network for controlling a gecko-like robot is adjusted based on the dung beetle-like robot model to accommodate the differences in the number of joints. 
Accordingly, the gecko-like robot with 16 DOFs is trained to walk on flat terrain over 500 generations.

\begin{figure}[t]
\centering
\includegraphics[width=0.4\textwidth]{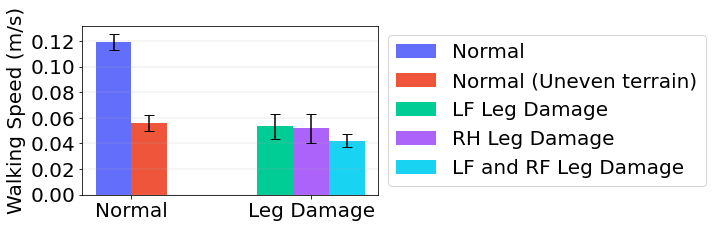}
\caption{Locomotion performance of a gecko-like robot, measured by walking speed, under various conditions: a normal robot with 16 DOFs on two different terrains (flat and uneven terrains), and a robot with three different types of morphological damage, including left front (LF), right hind (RH), and all front (LF and RF) leg damage.}
\label{fig:gecko_walking_speed}
\end{figure}

\begin{figure}[t]
\centering
\includegraphics[width=0.49\textwidth]{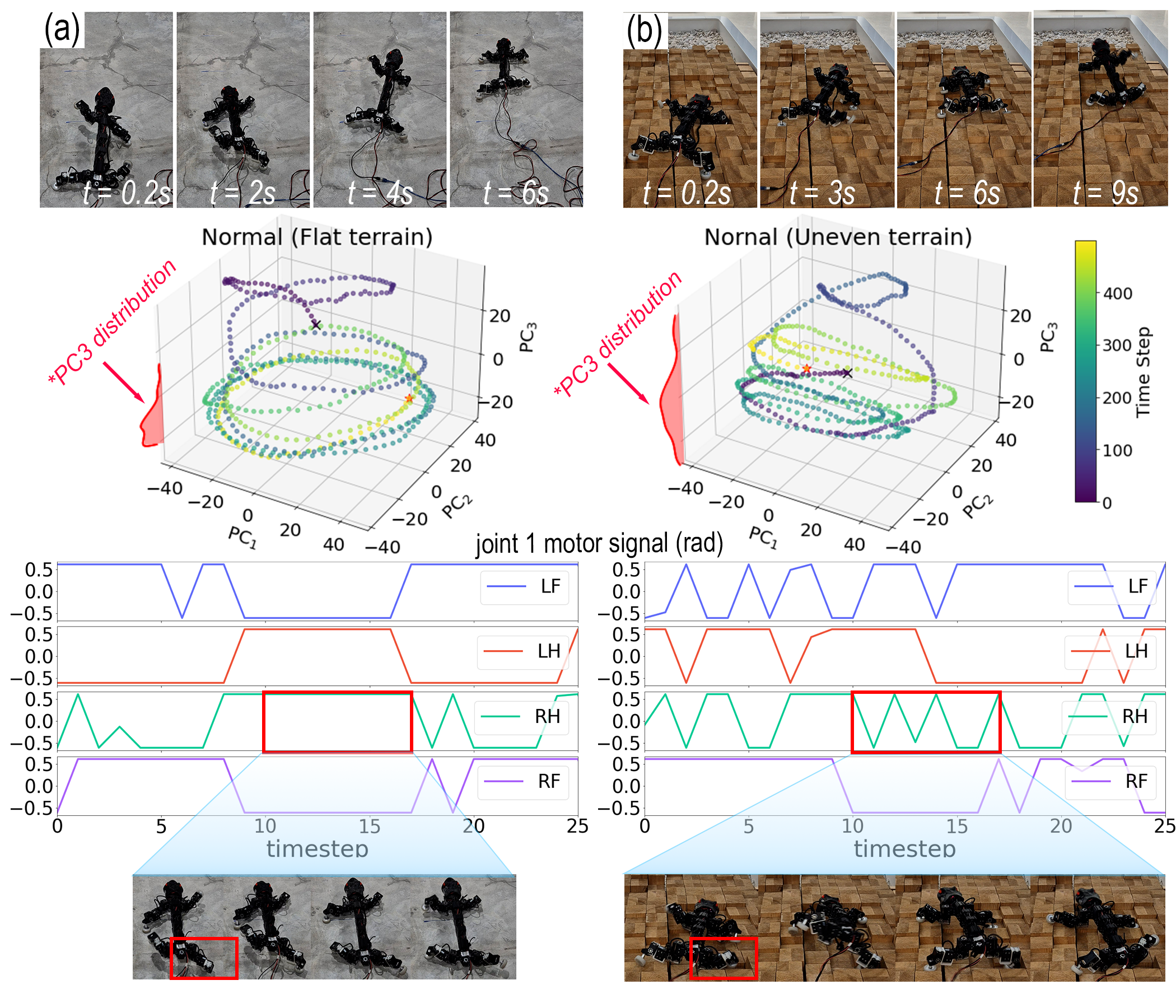}
\caption{The progression of gecko-like robot movements during forward walking on flat terrain (a) and uneven terrain (b), along with the PCA on the weight space and the corresponding changes in joint 1 (j1) trajectories over time. 
LF, LH, RH, and RF stand for left front, left hind, right hind, and right front, respectively. *PC$_3$ is the distribution of PC$_3$ after 100 time steps.
See video at \url{https://bit.ly/3D2ZHBf}.}
\label{fig:gecko_walking_legtraj}
\end{figure}

To evaluate the locomotion performance of the Hebbian network, five experiment results were recorded throughout the traversal of a one-meter distance.
As a result, the sim-to-real transfer demonstrates that the robot can achieve a speed of approximately 12 cm/s on flat terrain (Figs.~\ref{fig:gecko_walking_speed} and \ref{fig:gecko_walking_legtraj}a). 
% For verify Habbian network adaptability on unseen environment, we tested the robot's walking performance on uneven terrain.
To verify the adaptability of the Hebbian network in an unseen environment, we tested the robot's walking performance on uneven terrain, which consists of wooden blocks packed together, each with a base size of 10~cm x 10~cm but varying in height (Fig.~\ref{fig:gecko_walking_legtraj}b).
The results indicate that the robot can efficiently adapt, successfully traversing the complex terrain at a speed of approximately 6~cm/s (Fig.~\ref{fig:gecko_walking_speed}).
To analyze the adaptability of Hebbian plasticity in this scenario, we investigated the PCA on the weight space as well as the motor signals on the action space and compared them during walking on flat and uneven terrain.
The PCA trajectory shows different patterns between flat and uneven terrain, as shown in Figs.~\ref{fig:gecko_walking_legtraj}a and~\ref{fig:gecko_walking_legtraj}b, respectively.
On flat terrain, the trajectory remains compact and structured, primarily governed by PC$_1$ and PC$_2$, indicating a stable attractor, where weight updates follow a predictable pattern with minimal deviations.
In contrast, on uneven terrain, the trajectory becomes more dispersed and exploratory, suggesting that the network requires more dynamic weight adjustments to respond to environmental variations. This increased spread is driven by the greater engagement of PC$_3$, which plays a crucial role in enabling higher-dimensional adaptation. 
More specifically, the distribution of PC$_3$ after 100 time steps (see *PC$_3$ distribution in Fig.~\ref{fig:gecko_walking_legtraj}, middle) on uneven terrain is approximately twice as large as on flat terrain, with a standard deviation (SD) of 16.45 for uneven terrain and only 7.48 for flat terrain.
The stronger influence of PC$_3$ in uneven terrain suggests that the network leverages Hebbian plasticity to explore a broader range of online weight adaptation, enhancing its ability to adapt efficiently to unpredictable environments.
% Each joint's motor signals are shown in Fig.~\ref{fig:gecko_walking_legtraj}.
For motor signal analysis, when the robot traverses uneven terrain, the fluctuations in joint 1 (j1), responsible for forward/backward motion, occur more frequently compared to walking on flat terrain. This is highlighted by the red boundary in Figs.~\ref{fig:gecko_walking_legtraj}a and~\ref{fig:gecko_walking_legtraj}b (bottom).
Typically, the increased frequency of forward/backward motion in the leg can help a legged robot overcome obstacles, such as getting stuck in a hole or navigating rough terrain.
It is evident that Hebbian plasticity makes significant leg adaptations to the robot for walking and avoiding getting struck on uneven terrain.
% We also observed the different motor signals of j2, j3, and j4 compared between flat (Fig.~\ref{fig:gecko_walking_legtraj}a) and uneven terrain (Fig.~\ref{fig:gecko_walking_legtraj}b), as expected due to the terrain's different roughness.
Note that the motor signals in this case are not smooth. This is due to our emphasis on enabling rapid, unconstrained adaptability. However, smoother motor signals can be obtained by introducing a jerk penalty in the reward function \cite{lin2020anti}.

% Furthermore, three leg injury scenarios for walking on flat terrain were investigated to evaluate the locomotion performance of the Hebbian network in the presence of unseen morphologies.
% First, the left front (LF) leg damage, which implies that all joints of the LF leg were fixed at a constant default position (see Fig.~\ref{fig:gecko_morp_damage}).
% The second condition involves damage to the right hind (RH) leg, while the third condition affects both front legs.

\begin{figure}[t]
\centering
\includegraphics[width=0.48\textwidth]{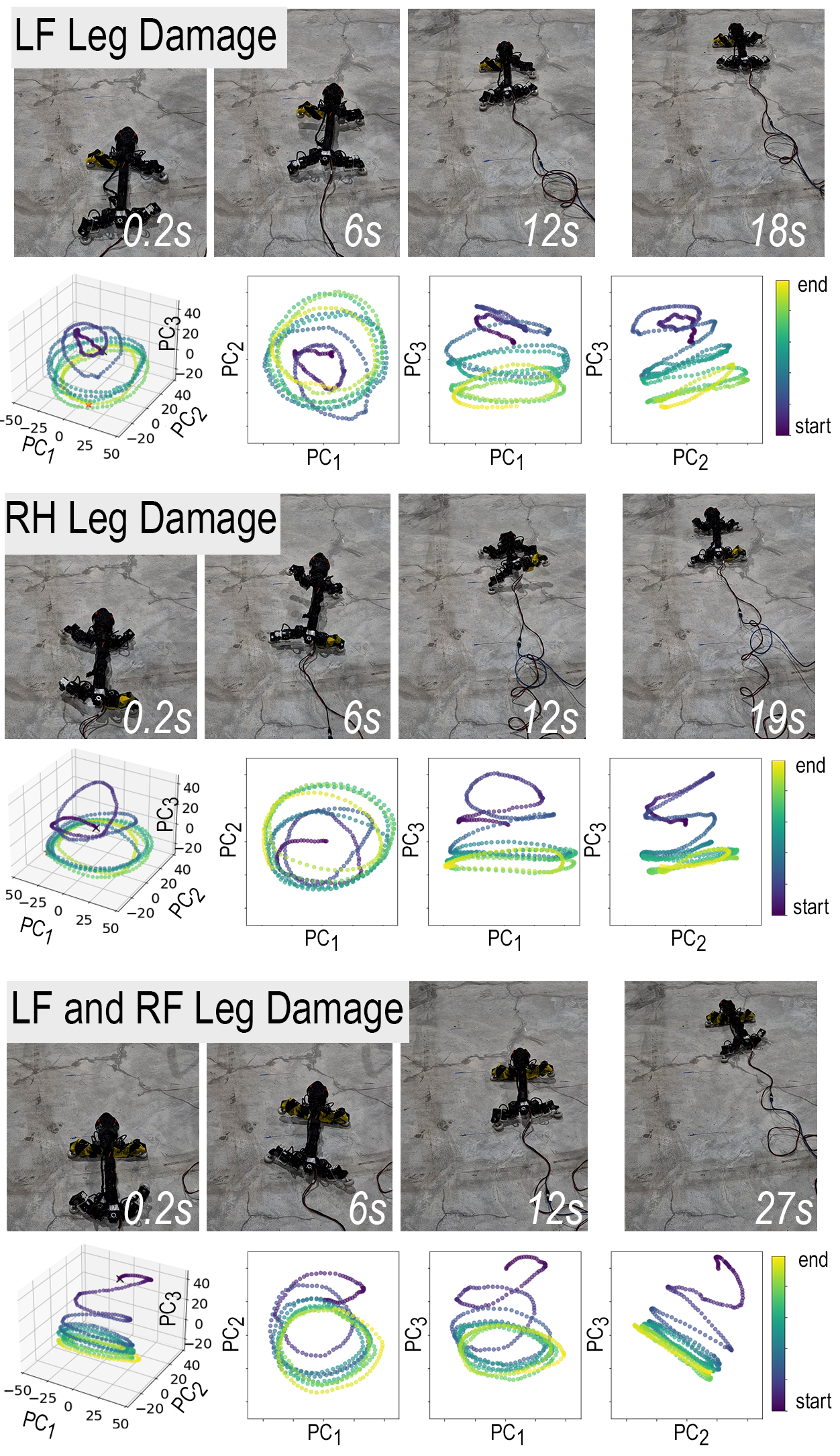}
\caption{The snapshots illustrate the robot's behavior for each type of morphological damage and PCA on weight space while walking on flat terrain. A video is available at \url{https://bit.ly/3D2ZHBf}.}
\label{fig:gecko_morp_damage}
\end{figure}

Furthermore, to evaluate the resilience of the Hebbian network under unseen morphological changes, three leg injury scenarios on flat terrain were investigated. First, the left front (LF) leg was damaged by fixing all its joints in a constant default position (Fig.~\ref{fig:gecko_morp_damage}, top). The second condition involved damage to the right hind (RH) leg (Fig.~\ref{fig:gecko_morp_damage}, middle), while the third condition affected both front legs (Fig.~\ref{fig:gecko_morp_damage}, bottom).
For brevity, these conditions will be referred to as LF damage, RH damage, and LF and RF damage, respectively.

The morphological damage results show that the robot can walk successfully at speeds of approximately 5 cm/s for LF damage, 5 cm/s for RH damage, and surprisingly 4 cm/s for LF and RF damage (Fig.~\ref{fig:gecko_walking_speed}).
Fig.~\ref{fig:gecko_morp_damage} illustrates the snapshots of the robot's posture over time for each type of morphological damage during walking, along with the corresponding PCA on the weight space observed from different views.
The PCA trajectories show that all morphological damage scenarios exhibit smooth attractors, indicating stable weight adaptation patterns. However, the PCA pattern for LF and RF leg damage differs slightly from the others, appearing more elliptical with a smaller spread along PC$_1$ and PC$_2$, whereas LF damage and RH damage exhibit a more circular shape.
This difference arises from the greater loss of sensory feedback, which constrains adaptation but still enables the robot to move forward effectively.
This suggests that Hebbian network not only enables locomotion adaptation but also ensures locomotion resilience effectively. It can deal with the loss of sensory feedback and controllable actuators.
In this study, the loss of proprioceptive feedback/control can be as high as 50\% of all legs, as demonstrated in the case of LF and RF leg damage.

\section{Discussion and Future Work}
\label{discussion}

% This shows that the Hebbian network can produce oscillations that not only take the foot contact into account.

%In the experiments, 
We trained and tested three policy variants (FF, Hebbian, and LSTM networks) for robot locomotion. The Hebbian network achieved similar performance to the other policies in training conditions. However, when testing the policies in OOD situations, the Hebbian network successfully enabled the real robot to walk, outperforming the other models in the sim-to-real transfer, where the feedforward and LSTM policies only enabled the robot to walk at a slow speed in non-straight lines (Fig.~\ref{real robot walk}).

%\begin{figure}[t]
%\centering
%\includegraphics[width=0.29\textwidth]{figures/objtrans_trainingcurve.pdf}
%\caption{Training curves: object transportation task. Parameter optimization with the ES in the object transportation task. The graph shows the reward over 500 generations of five trials on each model.}
%\label{train curve object}
%\end{figure}

We hypothesize that the superior performance of the Hebbian network in the OOD scenarios is due to a proportion of plastic parameters to the evolved parameters of the model. According to Table~\ref{tab:my-table_Number_parameters}, the Hebbian network has a similar number of evolved parameters to the LSTM network, but it has an order of magnitude more plastic parameters compared to the LSTM network. It has previously been speculated that a greater number of fixed parameters compared to plastic parameters can lead to a greater risk of overfitting \cite{kirsch2021meta}. 
% This might also help explain why the larger LSTM did not perform better than the smaller LSTM in OOD tasks.
For LSTM, the number of evolved parameters has a quadratic relationship with the number of plastic parameters, in contrast, this relationship is linear in the case of the Hebbian network.

The results from feedforward and LSTM policies point towards the overfitting to the specifics of the simulated robot in the training phase. One reason can be that the policies rely on the specific mass distribution, joint properties, and contact friction of the simulated robot, which may be too different from the real robot in the locomotion. The mass distribution in the simulated robot is more symmetrical compared to the real robot in both the left-to-right and head-to-rear directions. The difference in sim-to-real performance between the model types might have been smaller if the simulation was more faithful to the real world. However, creating realistic robot models for simulation is a high-effort task, highlighting the usefulness of the Hebbian network's increased generalization.

% \vspace{0.00mm}

% \begin{figure}[t]
% \centering
% \includegraphics[width=0.3\textwidth]{figures/PCA_lstm.png}
% % \caption{(b)}
% \caption{Trajectory of the principal component analysis (PCA) on the hidden states of the LSTM network on the locomotion task in robot simulation. Only the LSTM network with trained parameters exhibits a limit cycle in the PCs of the hidden states. }
% \label{PCA analysis LSTM}
% \end{figure}

% \vspace{0.00mm}

Interestingly, the Hebbian networks could adapt to the real-world robot without the randomization of terrain, mass distribution, joint properties, contact friction, and/or morphological damage in the training phase. The observed robustness of a policy with synaptic plasticity is in line with earlier research \cite{najarro2020, pedersen2021evolving}, however, these earlier studies demonstrated this robustness only on simulated robots.
Some randomization is intrinsically present in the Hebbian networks, i.e., the random initialization of the policy's weights at the beginning of an episode, which may contribute to more robust learning rules. However, the hidden states of the LSTM are also initialized randomly, and that did not prevent overfitting. 

Our study revealed an interesting dynamic of the Hebbian network, where foot contact inputs initiated walking behavior (Fig. \ref{fig:hebb_pick_up}). The oscillations of the principal components of the weights also stopped when the robot was lifted. The rapid changes of the synaptic strengths throughout the Hebbian network based on the foot contact inputs resulted in the non-oscillation and oscillation patterns of the principal components of the weights. This could be considered as synaptic self-organization of the Hebbian network caused by the Hebbian plasticity \cite{synaptic_self-organization}. To the best of our knowledge, such behavior has not been exhibited by other models in the literature.

We expanded our investigation of the Hebbian network's generalization capabilities by applying it to a different robot platform. A gecko-like robot with 16 DOFs was trained on flat terrain, and the learned policy was then transferred to a physical robot for real-world evaluation.
It was demonstrated that the gecko-like robot efficiently adapted its legs to traverse uneven terrain.
We observed a correlation between weight dynamics and action space underlying this adaptability. Specifically, the PCA trajectory of weight changes expanded along PC$_3$, enabling higher-dimensional adaptation, corresponding to higher-frequency fluctuations in the action space, particularly in joint 1.
Additionally, the Hebbian network effectively managed the loss of proprioceptive feedback/control, allowing the robot to achieve resilient locomotion dealing with leg damage.
As expected, greater leg loss imposed stronger constraints on adaptation. This was reflected in the PCA trajectory, where LF and RF leg damage produced an elliptical shape with a smaller spread along PC$_1$ and PC$_2$, while single-leg damage produced a more circular shape with greater spread.
% The robot was able to walk at 5 cm/s with LF or RH damage and at 4~cm/s with front damage.
% It was demonstrated that the gecko-like robot efficiently adapted its legs to traverse uneven terrain at a speed of 6~cm/s. Additionally, the Hebbian network effectively managed the loss of proprioceptive feedback/control, allowing the robot to maintain locomotion even with leg damage. The robot was able to walk at 5 cm/s with LF or RH damage and at 4~cm/s with front damage.

%Through visualization of the plastic synapses and the LSTM's hidden states using PCA (Figs. \ref{PCA analysis Hebb}, it can be observed that the principal components for both oscillate in stable patterns. 
%However, we did not observe the same behavior from the feedforward and LSTM networks on the real robot (Fig. \ref{fig:hebb_pick_up}), where the robot's legs always moved even there were no foot contact inputs. This could be explained by the fact that the joint position and IMU feedback might have made a greater contribution to the output of the LSTM network than the foot contact inputs.

In future studies, we will investigate whether combining randomization and Hebbian plasticity can lead to even more robust policies than any of the approaches alone. Recently, Di Guiro et al. investigated the gated recurrent unit (GRU), which, like Hebbian networks and LSTMs, has many plastic weights \cite{di2024meta}. They found that combining the GRU with domain randomization provided the best generalization to unseen robot morphologies. The Hebbian plasticity could also be combined with networks that can learn to grow their structure through the evolving process \cite{najarro2023towards}, thereby eliminating the need to predefine neural architecture.
% decide on the neural architecture a priori. 

In summary, our experiments demonstrate both the generalization of the Hebbian network to use in many robotic platforms and the abilities of synaptic plasticity to increase the generalization of the dynamic policy to OOD situations (sim-to-real adaptation, uneven terrain, and morphological damage) in both simulated and animal-like real-world robots. 
% While synaptic plasticity alone enables robust policies, there is ample opportunity to combine this approach with other approaches to continue to drive progress toward adaptive real-world robots.

%\begin{figure}[t]
%\centering
%\includegraphics[width=0.35\textwidth]{figures/norm.pdf}
%\caption{Comparison of using standard deviation and max normalization methods to normalize the dynamical weights of the Hebbian network. In both tasks, dividing with the maximum weight value results in the best-performing solution.}
%\label{compare normalization}

%\end{figure}

% Appendixes should appear before the acknowledgment.

% \section*{ACKNOWLEDGMENT}

%  ssssssssssssssssssssssssssssssssssssssssssssssssssssssssssssssssssssssssss ssssssssssssssssssssssssssssssssssssssssssssssssssssssssssssssssssssssssss ssssssssssssssssssssssssssssssssssssssssssssssssssssssssssssssssssssssssss.

%%%%%%%%%%%%%%%%%%%%%%%%%%%%%%%%%%%%%%%%%%%%%%%%%%%%%%%%%%%%%%%%%%%%%%%%%%%%%%%%

\bibliographystyle{IEEEtran}

\bibliography{IEEEfull}

\begin{thebibliography}{10}
\providecommand{\url}[1]{#1}
\csname url@rmstyle\endcsname
\providecommand{\newblock}{\relax}
\providecommand{\bibinfo}[2]{#2}
\providecommand\BIBentrySTDinterwordspacing{\spaceskip=0pt\relax}
\providecommand\BIBentryALTinterwordstretchfactor{4}
\providecommand\BIBentryALTinterwordspacing{\spaceskip=\fontdimen2\font plus
\BIBentryALTinterwordstretchfactor\fontdimen3\font minus
  \fontdimen4\font\relax}
\providecommand\BIBforeignlanguage[2]{{%
\expandafter\ifx\csname l@#1\endcsname\relax
\typeout{** WARNING: IEEEtran.bst: No hyphenation pattern has been}%
\typeout{** loaded for the language `#1'. Using the pattern for}%
\typeout{** the default language instead.}%
\else
\language=\csname l@#1\endcsname
\fi
#2}}

\bibitem{goodfellow2016deep}
I.~Goodfellow, Y.~Bengio, A.~Courville, and Y.~Bengio, \emph{Deep
  learning}.\hskip 1em plus 0.5em minus 0.4em\relax MIT Press, 2016, vol.~1.

\bibitem{vinyals2019grandmaster}
O.~Vinyals, \emph{et~al.}, ``Grandmaster level in starcraft ii using
  multi-agent reinforcement learning,'' \emph{Nature}, vol. 575, no. 7782, pp.
  350--354, 2019.

\bibitem{karoly2020deep}
A.~I. K{\'a}roly, P.~Galambos, J.~Kuti, and I.~J. Rudas, ``Deep learning in
  robotics: Survey on model structures and training strategies,'' \emph{IEEE
  Transactions on Systems, Man, and Cybernetics: Systems}, vol.~51, no.~1, pp.
  266--279, 2020.

\bibitem{mouha2021deep}
R.~A. Mouha \emph{et~al.}, ``Deep learning for robotics,'' \emph{Journal of
  Data Analysis and Information Processing}, vol.~9, no.~02, p.~63, 2021.

\bibitem{soori2023artificial}
M.~Soori, B.~Arezoo, and R.~Dastres, ``Artificial intelligence, machine
  learning and deep learning in advanced robotics, a review,'' \emph{Cognitive
  Robotics}, 2023.

\bibitem{heaven2019deep}
D.~Heaven \emph{et~al.}, ``Why deep-learning ais are so easy to fool,''
  \emph{Nature}, vol. 574, no. 7777, pp. 163--166, 2019.

\bibitem{zhang2018study}
C.~Zhang, O.~Vinyals, R.~Munos, and S.~Bengio, ``A study on overfitting in deep
  reinforcement learning,'' \emph{arXiv preprint arXiv:1804.06893}, 2018.

\bibitem{zhao2019investigating}
C.~Zhao, O.~Sigaud, F.~Stulp, and T.~M. Hospedales, ``Investigating
  generalisation in continuous deep reinforcement learning,'' \emph{arXiv
  preprint arXiv:1902.07015}, 2019.

\bibitem{zhao2020sim}
W.~Zhao, J.~P. Queralta, and T.~Westerlund, ``Sim-to-real transfer in deep
  reinforcement learning for robotics: a survey,'' in \emph{2020 IEEE symposium
  series on computational intelligence (SSCI)}.\hskip 1em plus 0.5em minus
  0.4em\relax IEEE, 2020, pp. 737--744.

\bibitem{Domain_randomization}
F.~Muratore, F.~Ramos, G.~Turk, W.~Yu, M.~Gienger, and J.~Peters, ``Robot
  learning from randomized simulations: A review,'' \emph{Frontiers in Robotics
  and AI}, vol.~9, 2022.

\bibitem{Self-Organized_Stick_Insect-Like_Locomotion}
A.~D. Larsen, \emph{et~al.}, ``Self-organized stick insect-like locomotion
  under decentralized adaptive neural control: From biological investigation to
  robot simulation (adv. theory simul. 8/2023),'' \emph{Advanced Theory and
  Simulations}, vol.~6, no.~8, p. 2370018, 2023.

\bibitem{Leung_Integrated_Modular}
B.~Leung, P.~Billeschou, and P.~Manoonpong, ``Integrated modular neural control
  for versatile locomotion and object transportation of a dung beetle-like
  robot,'' \emph{IEEE Transactions on Cybernetics}, pp. 1--14, 2023.

\bibitem{Citri2008}
\BIBentryALTinterwordspacing
A.~Citri and R.~C. Malenka, ``Synaptic plasticity: Multiple forms, functions,
  and mechanisms,'' \emph{Neuropsychopharmacology}, vol.~33, pp. 18--41, 2008.
  [Online]. Available: \url{https://doi.org/10.1038/sj.npp.1301559}
\BIBentrySTDinterwordspacing

\bibitem{pike1999postsynaptic}
F.~G. Pike, R.~M. Meredith, A.~W. Olding, and O.~Paulsen, ``Postsynaptic
  bursting is essential for ‘hebbian’induction of associative long-term
  potentiation at excitatory synapses in rat hippocampus,'' \emph{The Journal
  of physiology}, vol. 518, no.~2, pp. 571--576, 1999.

\bibitem{meliza2006receptive}
C.~D. Meliza and Y.~Dan, ``Receptive-field modification in rat visual cortex
  induced by paired visual stimulation and single-cell spiking,''
  \emph{Neuron}, vol.~49, no.~2, pp. 183--189, 2006.

\bibitem{cassenaer2007hebbian}
S.~Cassenaer and G.~Laurent, ``Hebbian stdp in mushroom bodies facilitates the
  synchronous flow of olfactory information in locusts,'' \emph{Nature}, vol.
  448, no. 7154, pp. 709--713, 2007.

\bibitem{najarro2020}
E.~Najarro and S.~Risi, ``Meta-learning through hebbian plasticity in random
  networks,'' \emph{Advances in Neural Information Processing Systems},
  vol.~33, 2020.

\bibitem{pedersen2021evolving}
J.~W. Pedersen and S.~Risi, ``Evolving and merging hebbian learning rules:
  increasing generalization by decreasing the number of rules,'' \emph{arXiv
  preprint arXiv:2104.07959}, 2021.

\bibitem{ferigo2021evolving}
A.~Ferigo, G.~Iacca, E.~Medvet, and F.~Pigozzi, ``Evolving hebbian learning
  rules in voxel-based soft robots.(2021),'' 2021.

\bibitem{chalvidal2022meta}
M.~Chalvidal, T.~Serre, and R.~Van-Rullen, ``Meta-reinforcement learning with
  self-modifying networks,'' in \emph{36th Conference on Neural Information
  Processing Systems (NeurIPS 2022)}, 2022, pp. 1--19.

\bibitem{palm2021testing}
R.~B. Palm, E.~Najarro, and S.~Risi, ``Testing the genomic bottleneck
  hypothesis in hebbian meta-learning,'' in \emph{NeurIPS 2020 Workshop on
  Pre-registration in Machine Learning}.\hskip 1em plus 0.5em minus 0.4em\relax
  PMLR, 2021, pp. 100--110.

\bibitem{Floreano2001}
D.~Floreano and J.~Urzelai, ``Evolution of plastic control networks,''
  \emph{Autonomous Robots}, vol.~11, no.~3, pp. 311--317, Nov 2001.

\bibitem{Billeschou2020_ALPHARobot}
P.~Billeschou, N.~N. Bijma, L.~B. Larsen, S.~N. Gorb, J.~C. Larsen, and
  P.~Manoonpong, ``Framework for developing bio-inspired morphologies for
  walking robots,'' \emph{Applied Sciences}, vol.~10, p. 6986, 2020.

\bibitem{haomachai2024transition}
W.~Haomachai, Z.~Dai, and P.~Manoonpong, ``Transition gradient from standing to
  traveling waves for energy-efficient slope climbing of a gecko-inspired
  robot,'' \emph{IEEE Robotics and Automation Letters}, 2024.

\bibitem{hochreiter1997long}
S.~Hochreiter and J.~Schmidhuber, ``Long short-term memory,'' \emph{Neural
  computation}, vol.~9, no.~8, pp. 1735--1780, 1997.

\bibitem{stanley2003evolving}
K.~O. Stanley, B.~D. Bryant, and R.~Miikkulainen, ``Evolving adaptive neural
  networks with and without adaptive synapses,'' in \emph{The 2003 Congress on
  Evolutionary Computation, 2003. CEC'03.}, vol.~4.\hskip 1em plus 0.5em minus
  0.4em\relax IEEE, 2003, pp. 2557--2564.

\bibitem{lstmsnake_Ouyang_2020}
S.~Ouyang and W.~Wei, ``Flexible adaptive control of snake-like robot based on
  lstm and gait,'' \emph{Journal of Physics: Conference Series}, vol. 1487,
  no.~1, p. 012049, mar 2020.

\bibitem{soltoggio2018born}
A.~Soltoggio, K.~O. Stanley, and S.~Risi, ``Born to learn: the inspiration,
  progress, and future of evolved plastic artificial neural networks,''
  \emph{Neural Networks}, vol. 108, pp. 48--67, 2018.

\bibitem{risi:book25}
\BIBentryALTinterwordspacing
S.~Risi, D.~Ha, Y.~Tang, and R.~Miikkulainen, \emph{Neuroevolution: Harnessing
  Creativity in AI Model Design}.\hskip 1em plus 0.5em minus 0.4em\relax
  Cambridge, MA: MIT Press, 2025. [Online]. Available:
  \url{http://www.cs.utexas.edu/users/ai-lab?risi:book25}
\BIBentrySTDinterwordspacing

\bibitem{soltoggio2008evolutionary}
A.~Soltoggio, J.~A. Bullinaria, C.~Mattiussi, P.~D{\"u}rr, and D.~Floreano,
  ``Evolutionary advantages of neuromodulated plasticity in dynamic,
  reward-based scenarios,'' in \emph{Proceedings of the 11th international
  conference on artificial life (Alife XI)}, no. CONF.\hskip 1em plus 0.5em
  minus 0.4em\relax MIT Press, 2008, pp. 569--576.

\bibitem{ben2020evolving}
E.~Ben-Iwhiwhu, P.~Ladosz, J.~Dick, W.-H. Chen, P.~Pilly, and A.~Soltoggio,
  ``Evolving inborn knowledge for fast adaptation in dynamic pomdp problems,''
  in \emph{Proceedings of the 2020 Genetic and Evolutionary Computation
  Conference}, 2020, pp. 280--288.

\bibitem{risi2010indirectly}
S.~Risi and K.~O. Stanley, ``Indirectly encoding neural plasticity as a pattern
  of local rules,'' in \emph{International Conference on Simulation of Adaptive
  Behavior}.\hskip 1em plus 0.5em minus 0.4em\relax Springer, 2010, pp.
  533--543.

\bibitem{cully2015robots}
A.~Cully, J.~Clune, D.~Tarapore, and J.-B. Mouret, ``Robots that can adapt like
  animals,'' \emph{Nature}, vol. 521, no. 7553, pp. 503--507, 2015.

\bibitem{MELA_quadruped}
C.~Yang, K.~Yuan, Q.~Zhu, W.~Yu, and Z.~Li, ``Multi-expert learning of adaptive
  legged locomotion,'' \emph{Science Robotics}, vol.~5, no.~49, p. eabb2174,
  2020.

\bibitem{kumar2021rma}
A.~Kumar, Z.~Fu, D.~Pathak, and J.~Malik, ``Rma: Rapid motor adaptation for
  legged robots,'' \emph{arXiv preprint arXiv:2107.04034}, 2021.

\bibitem{wang2023learning}
Y.~Wang, Z.~Jiang, and J.~Chen, ``Learning robust, agile, natural legged
  locomotion skills in the wild,'' in \emph{RoboLetics: Workshop on Robot
  Learning in Athletics@ CoRL 2023}, 2023.

\bibitem{zhang2018mode}
H.~Zhang, S.~Starke, T.~Komura, and J.~Saito, ``Mode-adaptive neural networks
  for quadruped motion control,'' \emph{ACM Transactions on Graphics (TOG)},
  vol.~37, no.~4, pp. 1--11, 2018.

\bibitem{tobin2017domain}
J.~Tobin, R.~Fong, A.~Ray, J.~Schneider, W.~Zaremba, and P.~Abbeel, ``Domain
  randomization for transferring deep neural networks from simulation to the
  real world,'' in \emph{2017 IEEE/RSJ international conference on intelligent
  robots and systems (IROS)}.\hskip 1em plus 0.5em minus 0.4em\relax IEEE,
  2017, pp. 23--30.

\bibitem{wang2022generalizing}
J.~Wang, \emph{et~al.}, ``Generalizing to unseen domains: A survey on domain
  generalization,'' \emph{IEEE Transactions on Knowledge and Data Engineering},
  2022.

\bibitem{muratore2022robot}
F.~Muratore, F.~Ramos, G.~Turk, W.~Yu, M.~Gienger, and J.~Peters, ``Robot
  learning from randomized simulations: A review,'' \emph{Frontiers in Robotics
  and AI}, p.~31, 2022.

\bibitem{peng2018sim}
X.~B. Peng, M.~Andrychowicz, W.~Zaremba, and P.~Abbeel, ``Sim-to-real transfer
  of robotic control with dynamics randomization,'' in \emph{2018 IEEE
  international conference on robotics and automation (ICRA)}.\hskip 1em plus
  0.5em minus 0.4em\relax IEEE, 2018, pp. 3803--3810.

\bibitem{akkaya2019solving}
I.~Akkaya, \emph{et~al.}, ``Solving rubik's cube with a robot hand,''
  \emph{arXiv preprint arXiv:1910.07113}, 2019.

\bibitem{DBLP:conf/corl/RudinHR021}
N.~Rudin, D.~Hoeller, P.~Reist, and M.~Hutter, ``Learning to walk in minutes
  using massively parallel deep reinforcement learning,'' in \emph{Conference
  on Robot Learning, 8-11 November 2021, London, {UK}}, ser. Proceedings of
  Machine Learning Research, A.~Faust, D.~Hsu, and G.~Neumann, Eds., vol.
  164.\hskip 1em plus 0.5em minus 0.4em\relax {PMLR}, 2021, pp. 91--100.

\bibitem{salimans2017evolution}
T.~Salimans, J.~Ho, X.~Chen, S.~Sidor, and I.~Sutskever, ``Evolution strategies
  as a scalable alternative to reinforcement learning,'' \emph{arXiv preprint
  arXiv:1703.03864}, 2017.

\bibitem{Efficacy_of_Modern_Neuro-Evolutionary}
P.~Pagliuca, N.~Milano, and S.~Nolfi, ``Efficacy of modern neuro-evolutionary
  strategies for continuous control optimization,'' \emph{Frontiers in Robotics
  and AI}, vol.~7, 2020.

\bibitem{lin2020anti}
Y.~Lin, J.~McPhee, and N.~L. Azad, ``Anti-jerk on-ramp merging using deep
  reinforcement learning,'' in \emph{2020 IEEE Intelligent Vehicles Symposium
  (IV)}.\hskip 1em plus 0.5em minus 0.4em\relax IEEE, 2020, pp. 7--14.

\bibitem{kirsch2021meta}
L.~Kirsch and J.~Schmidhuber, ``Meta learning backpropagation and improving
  it,'' \emph{Advances in Neural Information Processing Systems}, vol.~34, pp.
  14\,122--14\,134, 2021.

\bibitem{synaptic_self-organization}
M.~Dehghani-Habibabadi and K.~Pawelzik, ``Synaptic self-organization of
  spatio-temporal pattern selectivity,'' \emph{PLOS Computational Biology},
  vol.~19, no.~2, pp. 1--26, 02 2023.

\bibitem{di2024meta}
F.~Di~Giuro, F.~Zargarbashi, J.~Cheng, D.~Kang, B.~Sukhija, and S.~Coros,
  ``Meta-reinforcement learning for universal quadrupedal locomotion control,''
  \emph{arXiv preprint arXiv:2407.17502}, 2024.

\bibitem{najarro2023towards}
E.~Najarro, S.~Sudhakaran, and S.~Risi, ``Towards self-assembling artificial
  neural networks through neural developmental programs,'' in \emph{Artificial
  Life Conference Proceedings 35}, vol. 2023, no.~1.\hskip 1em plus 0.5em minus
  0.4em\relax MIT Press, 2023, p.~80.

\end{thebibliography}

% \begin{thebibliography}{99}
% \end{thebibliography}

\end{document}